# RFWNet: A Lightweight Remote Sensing Object Detector Integrating Multi-Scale Receptive Fields and Foreground Focus Mechanism

Yujie Lei, Wenjie Sun, *Student Member, IEEE*, Sen Jia, *Senior Member, IEEE*, Qingquan Li, Jie Zhang

***Abstract*—Challenges in remote sensing object detection (RSOD), such as high inter-class similarity, imbalanced foreground-background distribution, and the small size of objects in remote sensing images significantly hinder detection accuracy. Moreover, the trade-off between model accuracy and computational complexity poses additional constraints on the application of RSOD algorithms. To address these issues, this study proposes an efficient and lightweight RSOD algorithm integrating multi-scale receptive fields and foreground focus mechanism, named Robust Foreground Weighted Network (RFWNet). Specifically, we proposed a lightweight backbone network Receptive Field Adaptive Selection Network (RFASNet), leveraging the rich context information of remote sensing images to enhance class separability. Additionally, we developed a Foreground Background Separation Module (FBSM) consisting of a Background Redundant Information Filtering Module (BRIFM) and a Foreground Information Enhancement Module (FIEM) to emphasize critical regions within images while filtering redundant background information. Finally, we designed a loss function, the Weighted CIoU-Wasserstein loss ($L_{WCW}$), which weights the IoU-based loss by using the Normalized Wasserstein Distance to mitigate model sensitivity to small object position deviations. The comprehensive experimental results demonstrate that RFWNet achieved 95.3% and 73.2% mAP with 6.0M parameters on the DOTA V1.0 and NWPU VHR-10 datasets, respectively, with an inference speed of 52 FPS.**

This work was supported by Major Science and Technology Special Program of Yunnan Province (202202AE090034), the National Key Research and Development Program (2022YFF0713002, 2021YFD1901001), and Project (GEMLab-2023001) supported by MNR Key Laboratory for Geo-Environmental Monitoring of Greater Bay Area. (Corresponding author: Jie Zhang.)

Yujie Lei, Jie Zhang, and Wenjie Sun are with College of Information and Electrical Engineering, China Agricultural University, Beijing 100083, China, Key Laboratory of Agricultural Machinery Monitoring and Big Data Applications, Ministry of Agriculture and Rural Affairs, Beijing 100083, China and National Innovation Center for Digital Agricultural Products Circulation, China Agricultural University, Beijing 100083, China. (e-mail: lyj_ciee@cau.edu.cn, jiezhang@cau.edu.cn, swj@cau.edu.cn).

Sen Jia is with the College of Computer Science and Software Engineering, Shenzhen University, Shenzhen 518060, China (e-mail: senjia@szu.edu.cn).

Qingquan Li is with Guangdong Key Laboratory of Urban Informatics, Guangdong Laboratory of Artificial Intelligence and Digital Economy (SZ), Shenzhen Key Laboratory of Spatial Information Smart Sensing and Services, and Research Institute of Smart Cities, Shenzhen University, Shenzhen 518060, China (e-mail: liqq@szu.edu.cn).

## I. INTRODUCTION

RAPID development of deep learning has significantly enhanced computer vision technology, particularly in the field of object detection, which has wide application potential in automatic driving, environmental monitoring, and smart agriculture [1]. With the progress in airborne and satellite sensors, high spatial resolution (HSR) remote sensing images have become increasingly accessible. Therefore, RSOD based on deep learning has gained prominence, playing a critical role in various domains, including ecological monitoring and smart agriculture [2].

However, remote sensing objects are complex and variable in scale, features, and distribution, which makes RSOD face challenges such as high inter-class similarity, small target size, and imbalance between foreground and background distribution [3]. High inter-class similarity can lead to misclassification without referring to sufficient contextual information. Additionally, the proportion of background and small target size further complicate accurate detection [4].

To address these challenges, many high-precision RSOD algorithms have been proposed [5]. For instance, SOD-YOLO improves the detection of small targets by fusing local and global information, but it does not fully consider the issue of imbalanced foreground-background distribution [6]. Lv et al. proposed a detection model tailored for small objects, which effectively mitigates the small-target detection challenge [7], yet it also suffers from the problem that the model tends to predict targets as background. Rao et al. proposed a cross-grid label assignment to enhance positive samples to alleviate the unbalanced foreground-background distribution issue, but the accuracy in detecting small and high interclass similarity targets was limited by limited exploration of context information [8]. Additionally, most of the existing research on RSOD primarily focuses on improving accuracy while neglecting computational complexity, leading to methods with high accuracy but excessive model parameters and slow inference speed [9]. With the development of UAVs across various fields, the demand for real-time RSOD algorithms has intensified, and the balance between model accuracy and speed has become increasingly critical. Consequently, developing lightweight models that can effectively address the challenges of high inter-class similarity, imbalanced foreground-background distribution, and the small size of objects in RSOD has become the key focus in this area.

To address the challenges of high inter-class similarity,

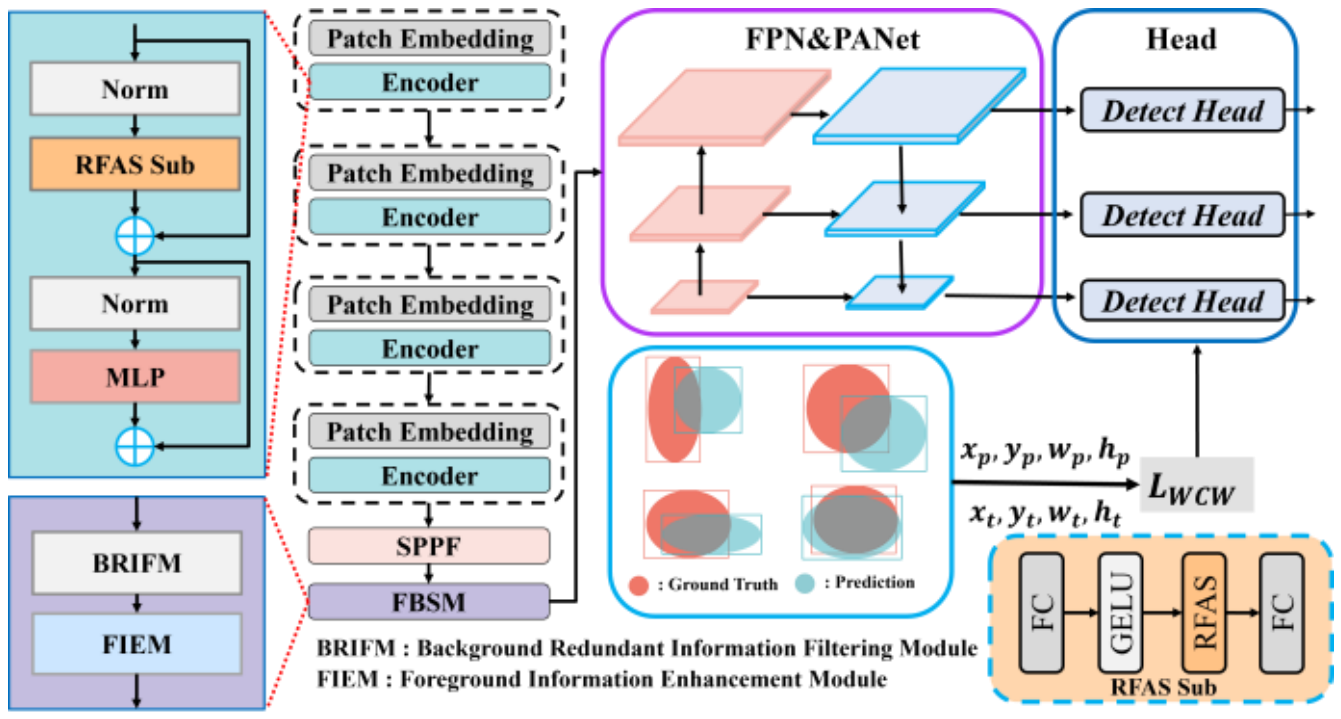

**Fig. 1.** RFWNet structure diagram.

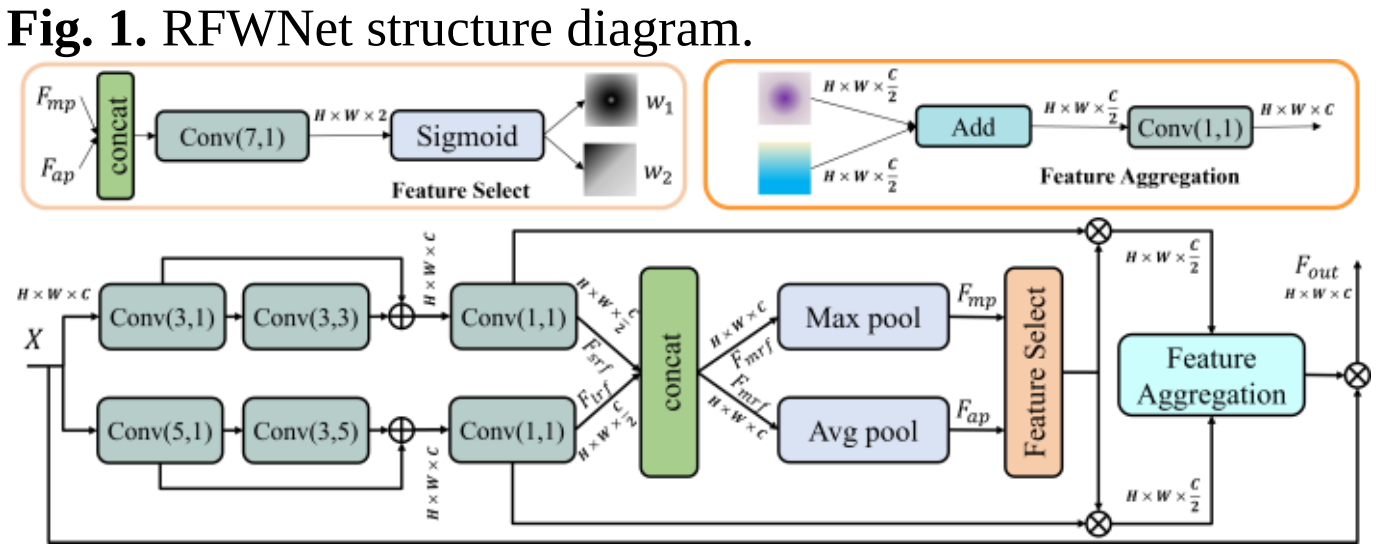

**Fig. 2.** RFAS Block structure diagram. Corresponding to RFAS in RFAS Sub in Fig.1.

small target size, and unbalanced foreground-background distribution in RSOD, while keeping a balance between accuracy and model complexity, we propose a lightweight and high-precision RSOD model, RFWNet. The principal contributions of this study are outlined as follows:

1) We design a lightweight backbone Receptive Field Adaptive Selection Network (RFASNet) as the backbone based on convolutions with different expansion rates. It is designed to fully utilize the context information of targets based on adaptive selection of receptive fields, thereby increasing the interclass discrete distance.
2) We develop a Foreground Background Separation Module comprising a background redundant information filtering module and a foreground information enhancement module to extract more effective features while filtering out redundancy.
3) We propose a novel regression loss function, Weighted CIoU-Wasserstein loss ($L_{WCW}$), which weights the IoU-based loss by using the Normalized Wasserstein Distance to mitigate the model's sensitivity to small target positional deviations.

## II. Methodology

RFWNet consists of backbone, neck and head. The backbone consists of RFASNet, Spatial Pyramid Pooling-Fast and FBSM. Neck consists of Feature Pyramid Network (FPN) and Path Aggregation Network (PANet) [10]. Head consists of three detection heads. The network is shown in Fig. 1.

### *A. Receptive Field Adaptive Selection Network (RFASNet)*

Different sizes of receptive fields are accompanied by varying target context information [11]. The rich contextual information provided by large receptive fields offers a significant advantage in recognizing confusable classes and large area features. However, large receptive fields are not always necessary. For targets with distinct local features, a smaller receptive field is often more effective. To address the challenge of inter-class similarity, we propose RFASNet as backbone. This network is designed to adaptively select the appropriate receptive field size based on the characteristics of targets. As shown in Fig.1, RFASNet consists of Receptive Field Adaptive Selection Sub block (RFAS Sub), MLP, and normalization. The core of RFAS Sub is RFAS Block.

Specifically, RFAS Block consists of three parts: multiple different receptive field branches, feature selection and feature aggregation. Multiple different receptive field branches are composed of multi-scale convolutions, and these layers can extract features with different receptive fields from the input image to provide rich context information. Since large receptive fields are often realized by large kernel convolution, but large kernel convolution often brings large computation, we decompose the large kernel convolution into standard convolution and dilated convolution to reduce the model computation while guaranteeing large receptive fields. The kernel size (k) and dilation rate (d) are crucial parameters for RFAS Block. Extensive experiments were conducted to explore their combinations under the constraint of a fixed small receptive field branch. The optimal performance was achieved with two consecutive convolutions in the large receptive field branch configured as (k=5, d=1) and (k=3, d=5). The structure of RFAS Block as shown in Fig. 2.

We superimpose the features of two smaller receptive fields and the features of two larger receptive fields to obtain the small receptive field feature ($\boldsymbol{F_{srf}}$) and the large receptive field feature ($\boldsymbol{F_{lrf}}$), respectively. Then the Concat operation is used to splice $F_{srf}$ and $F_{lrf}$ in the channel dimension to obtain the mixed receptive fields feature ($\boldsymbol{F_{mrf}}$).

$$F_{srf} = W_{3,3}\left(W_{3,1}(x)\right) + W_{3,1}(x) \tag{1}$$

$$F_{lrf} = W_{3,5}\left(W_{5,1}(x)\right) + W_{5,1}(x) \tag{2}$$

where $\boldsymbol{W_{a,b}}$ denotes the convolution operation, $a$ denotes the kernel size, $\boldsymbol{b}$ denotes the dilation rate, $\boldsymbol{X}$ denotes the input.

To select the most suitable receptive field for the target, we perform pooling operations on multi receptive field mixed features $\boldsymbol{F_{mrf}}$ to highlight key information. The outputs of maximum pooling and average pooling $\boldsymbol{F_{mp}}$ and $\boldsymbol{F_{ap}}$ are concatenated along the channel dimension and used as input to the feature selection module. This module first generates a spatial attention map by convolution operation to encode the importance information of each spatial location. Subsequently, the attention map is mapped into spatial selection weights ($\boldsymbol{w}$) using sigmoid ($\sigma$) function to quantify the dependency of each spatial location on different receptive field features.

$$w = \sigma\left(W_{7,1}([F_{mp}, F_{ap}])\right) \tag{3}$$

where the middle bracket denotes concat.

Based on the 2-channel weight $\boldsymbol{w}$, the feature aggregation module aggregates $\boldsymbol{F_{srf}}$ and $\boldsymbol{F_{lrf}}$ from different receptive field branches by weighted summation operation to obtain the final attention feature map. Finally, the attention feature map is multiplied by the input $\boldsymbol{X}$ to enhance the input feature map.

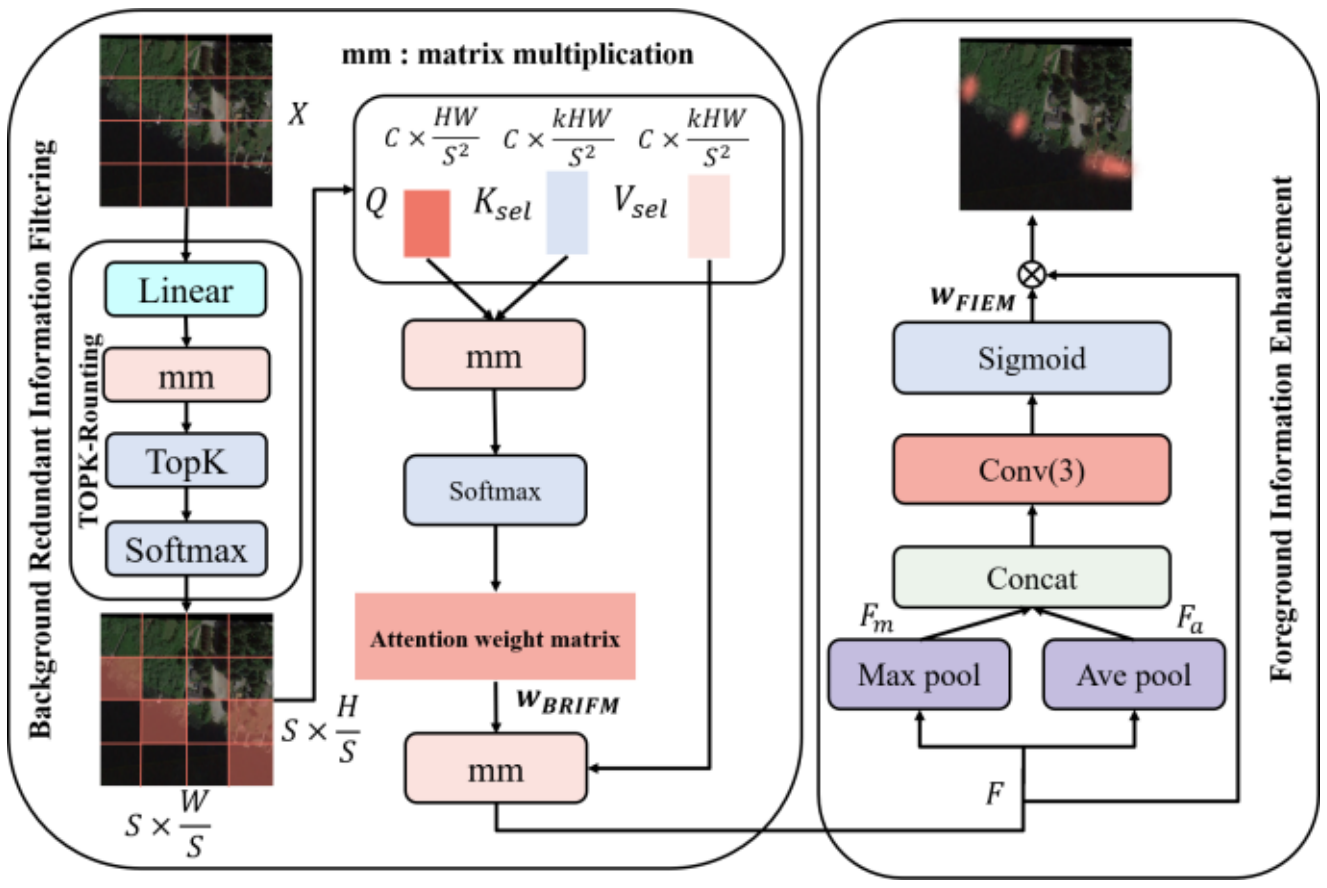

**Fig. 3.** FBSM structure diagram.

$$F_{out} = x * W_{1,1}(w_1 * F_{srf} + w_2 * F_{lrf}) \quad (4)$$

*B. Foreground Background Separation Module (FBSM)*

In remote sensing images, complex background often occupies a large proportion, while the foreground proportion is small [12]. To filter redundant background information and enhance the focus on key information, we design the FBSM, as shown in Fig. 3. FBSM includes two parts: Background Redundant Information Filtering Module (BRIFM) and Foreground Information Enhancement Module (FIEM). Firstly, BRIFM divides the image into multiple regions, and then filters out the irrelevant regions according to the key-value pairs to filter the redundant information. Secondly, FIEM dynamically adjusts the importance of each pixel on the result of BRIFM, and enhances the model's attention to key areas.

Specifically, BRIFM mainly consists of linear transformation and region filtering. For a given input feature map $\boldsymbol{X} \in \mathbb{R}^{H\times W\times C}$, it is first divided into multiple non-overlapping regions containing $\frac{H\times W}{S^2}$ feature vectors, then $\boldsymbol{X}$ can be reshape as $\boldsymbol{X_r} \in \mathbb{R}^{S^2\times\frac{H\times W}{S^2}\times C}$. Next, the query tensor $\boldsymbol{Q}$, the key tensor $\boldsymbol{K}$, and the value tensor $\boldsymbol{V}$, are obtained by linear transformation mapping, and the mean value of each tensor in each region is computed to capture the overall features of each region. Subsequently, use TopkRouting to select the most relevant K key-value pairs, return the routing weight $\boldsymbol{r_w}$ and index $r_I$, and assemble the key-value pairs $\boldsymbol{K_{sel}}$ and $\boldsymbol{V_{sel}}$ based on the indexes and weights. Then input the dot product of $\boldsymbol{Q}$ and $\boldsymbol{K_{sel}}$ into the Softmax function to get the attention weight matrix ($\boldsymbol{w_{BRIFM}}$). Finally, the filtering of irrelevant region is realized by multiplying $\boldsymbol{V_{sel}}$ with $\boldsymbol{w_{BRIFM}}$.

The FIEM is based on the results of the BRIFM, which mainly consists of pooling, convolutional and Sigmoid layers. Specifically, the feature $\boldsymbol{F}$ processed by BRIFM are used as input for max pooling and average pooling to efficiently extract spatial relations. The outputs of max pooling and average pooling are $\boldsymbol{F_m}$ and $\boldsymbol{F_a}$ respectively. Then $\boldsymbol{F_m}$ and $\boldsymbol{F_a}$ are converted into a single-channel spatial attention map by a convolutional layer. Finally, the attention map is converted to the foreground information attention weight ($\boldsymbol{w_{FIEM}}$) using a sigmoid function, which is applied to the input feature map $\boldsymbol{F}$ to enhance the features of important regions.

*C. Weighted CIoU-Wasserstein loss*

Accurately detecting large number of small remote sensing targets is one of the main challenges of RSOD [13]. Limited by the calculation formula of the IoU itself, traditional IoU-based measurement methods are susceptible to the positional deviation of small objects. The IoU drops significantly when the small target has only a small positional deviation, this is unfriendly to RSOD. To address the problem, we propose a Weighted CIoU-Wasserstein loss ($L_{WCW}$).

Normalized Wasserstein Distance loss (NWD loss) is based on Wasserstein distance for metrics and is specifically designed to solve the problem of small target detection [14]. Firstly, the predicted boxes and labels are transformed into 2D Gaussian distributions to describe the weight distribution of different pixels in the bounding box. Secondly, NWD Loss measures the similarity between the predicted bounding box and the real bounding box by considering their difference in probability distribution. NWD loss not only relies on the spatial correlation between pixels, but also takes into account the location and size information of the bounding box, which is scale-insensitive and friendly for small target detection. Therefore, we further optimize the model's sensitivity to small target positional deviation by weighted combining NWD Loss based on CIoU, which is calculated as shown in Eq. (5).

$$L_{WCW} = \gamma * L_{CIoU} + \beta * L_{nwd} \quad (5)$$

where $\boldsymbol{L_{CIoU}}$ is the CIoU loss, the calculation process is shown in Eq. (6). $\boldsymbol{L_{nwd}}$ is the NWD loss, the calculation process is shown in Eq. (7-8). $\boldsymbol{\gamma}$ and $\boldsymbol{\beta}$ are the weights of $\boldsymbol{L_{CIoU}}$ and $\boldsymbol{L_{nwd}}$, and both of them are 0.5 by default.

$$L_{CIoU} = 1 - IoU + \frac{d^2}{c^2} + \frac{v^2}{(1-IoU)+v} \quad (6)$$

$$W_2^2(N_a, N_b) = \left\| \left( \left[c_{xa}, c_{ya}, \frac{W_a}{2}, \frac{H_a}{2}\right]^T, \left[c_{xb}, c_{yb}, \frac{W_b}{2}, \frac{H_b}{2}\right]^T \right) \right\|_2^2 \quad (7)$$

$$L_{nwd} = 1 - e^{-\frac{\sqrt{W_2^2(N_a,N_b)}}{C}} \quad (8)$$

where $\boldsymbol{d}$ represents the distance between the centroid of the prediction box and the centroid of the label, $\boldsymbol{v}$ is the correction factor, $\boldsymbol{c}$ represents the diagonal distance of the smallest outer rectangle of the two rectangular boxes, $(c_{xa}, c_{ya}, W_a, H_a)$ and $(c_{xb}, c_{yb}, W_b, H_b)$ are prediction boxes and real labels.

## III. Experiments and Discussion

*A. Dataset*

The DOTA V1.0 [15] includes 2806 images, 15 categories. We refer to the official setting to crop these images to images with a resolution of 1024×1024. The overlap between neighboring slices is 200. After cropping, the numbers of train, val, and test are 15749, 5297, and 10833, respectively.
The NWPU VHR-10 includes 800 images, 10 categories [16], divided in an 8:1:1 ratio.

*B. Implementation Detail and Evaluation Metrics*

**Implementation Detail.** All experiments in this study were conducted on the Ubuntu system based on the Pytorch framework, with Python version 3.8, and Cuda version 11.3. The graphics card is RTX3090. The initial learning rate is 0.001, with 16 batch sizes and 100 epochs. None of the experiments use pretraining weights. $\gamma = 0.5$, $\beta = 0.5$.

TABLE I
COMPARISON EXPERIMENT RESULTS OF NWPU VHR-10. RED INDICATES OPTIMAL, BLUE SUB-OPTIMAL.

| Methods | AE | SP | ST | BD | TC | BC | GT | HR | BE | VE | mAP (%) | Para(M) |
|---|---|---|---|---|---|---|---|---|---|---|---|---|
| F-RCNN [17] | 82.8 | 77.5 | 52.5 | 96.3 | 62.9 | 68.8 | 98.4 | 82.5 | 78.8 | 63.8 | 76.4 | 41.1 |
| M-RCNN [18] | 93.2 | 75.5 | 92.9 | 90.4 | 90.3 | 91.2 | 95.2 | 75.2 | 60.6 | 74.2 | 83.9 | 44.7 |
| YOLO-PDNet [10] | 99.0 | 88.2 | **97.7** | 97.2 | 94.2 | 93.4 | **99.5** | 90.0 | 83.3 | **93.5** | 93.6 | 68.2 |
| MEDNet [2] | 99.2 | 94.4 | 82.2 | **98.5** | **95.4** | 95.2 | 98.3 | 88.1 | 75.1 | 89.3 | 91.6 | - |
| SOD-YOLOv10 [6] | 99.5 | **99.2** | 90.1 | 96.7 | 91.4 | 90.6 | 96.5 | **95.8** | **92.0** | 72.8 | 93.2 | 28.0 |
| MashFormer [19] | **100** | 93.4 | 92.6 | 96.9 | 94.6 | **95.6** | **100** | 79.5 | **94.1** | 91.3 | 93.8 | - |
| RS-YOLO [9] | **99.9** | 94.1 | **96.6** | 97.8 | 93.8 | 92.7 | 97.6 | 89.5 | 84.7 | 92.3 | **93.9** | 31.0 |
| Baseline | 98.4 | 91.3 | 82.8 | 97.9 | 90.5 | 76.8 | 96.5 | 89.3 | 83.6 | 78.4 | 88.9 | **7.1** |
| ours | 99.8 | **95.3** | 83.0 | **97.9** | **98.9** | **98.7** | 99.4 | **95.7** | 90.3 | **94.2** | **95.3** | **6.0** |

TABLE II
COMPARISON EXPERIMENT RESULTS OF DOTA V1.0

| Methods | FMSSD [20] | Baseline | RetinaNet | ICN [21] | YOLOv7 [22] | IoU-Ada [5] | F-RCNN | RFWNet |
|---|---|---|---|---|---|---|---|---|
| mAP | **72.4** | 69.0 | 67.5 | 68.3 | 72.0 | 71.0 | 70.8 | **73.2** |
| Parameter(M) | - | **7.1** | 36.1 | - | 71.4 | - | 41.1 | **6.0** |

**Evaluation Metrics.** Average Precision (AP), Precision, Recall and mean Average Precision (mAP) are widely used evaluation indicators for RSOD, and they are also selected as experimental evaluation indicators in this study.

*C. Comparison Experiment*

To evaluate the performance of the RFWNet model more comprehensively, we designed a comparison experiment between RFWNet and mainstream lightweight RSOD model based on DOTA V1.0 and NWPU VHR-10 datasets.

**Experimental results on the NWPU VHR-10 Dataset.** To verify the effectiveness of our proposed method, comparative experiments are conducted on the NWPU VHR-10 dataset. The Table I shows that RFASNet has minimal parameters at 6.0M and achieves advanced results with a mAP of 95.3%. In terms of individual category performance, RFWNet achieves optimal or sub-optimal accuracy on 5 categories of NWPU VHR-10 and outperforms the baseline in all categories. Especially on the typical small-target category VE, RFWNet improves AP by 15.8% compared to the baseline. This demonstrates that RFWNet is effective in mitigating the effects of high inter-class similarity and small targets. However, since multi-scale receptive fields based on standard square convolution inevitably introduce some redundant noise, this limits the improvement magnitude for large aspect ratio targets. The qualitative results in Fig. 4 show that our RFWNet can effectively reduce wrong and missed detections caused by small targets and background interference.

**Experimental results on the DOTA V1.0 Dataset.** To verify the robustness of RFWNet, comparative experiments are conducted on the DOTA V1.0 dataset. Table II shows that RFWNet has minimal parameters at 6.0M and achieves advanced results with a mAP of 73.2% on this benchmark dataset. The qualitative results in Fig. 4 show that RFWNet has better detection ability for remote sensing targets of different scales in different scenes.

**Inference speed.** To better evaluate the lightweight of RFWNet, we conducted inference speed test experiments on the testset of DOTA V1.0. The results show that the RFWNet inference of a single image takes only 19.1 ms, achieving a FPS of 52. It can be seen that RFWNet has an excellent performance in both parameters and inference speed.

*D. Ablation Study*

**The influence of $L_{WCW}$'s hyperparameters.** In this section, we discuss the effect of the two hyperparameters $\gamma$ and $\beta$ on the accuracy. As shown in the experimental results in Table III. the best performance of 95.3% can be achieved when both $\gamma$ and $\beta$ are 0.5. We believe that this ratio can effectively reduce the sensitivity to small target position deviation while ensuring the effectiveness of bounding box regression for medium and large targets. Therefore, we used 0.5 as the default value of $\gamma$ and $\beta$ in all the experiments.

**Effectiveness of RFASNet.** In this section, we discuss the effectiveness of RFASNet. Table IV shows that compared to the baseline, our proposed backbone RFASNet brings 3.2% and 3.9% accuracy improvement on the two datasets while reducing the model parameters by 1.4M. The experimental results illustrate that RFASNet can accurately consider the target features and effectively solve the misclassification problem caused by high inter-class similarity. Therefore, we use RFASNet as backbone in the subsequent experiments.

**Effectiveness of FBSM.** In this section, we discuss the effectiveness of FBSM. Table IV shows that compared to RFASNet, FBSM brings 0.8% and 0.9% improvement in mAP on the two datasets, respectively. In addition, FBSM can bring gains of 0.7% and 0.9% when combined with the other two improvements. Although it brings a parameter improvement of 0.2M, this is acceptable compared to the improved mAP. The above experimental results illustrate that FBSM can filter the redundant background information and improve the focus on the foreground during the training process, effectively solving the foreground-background imbalance problem.

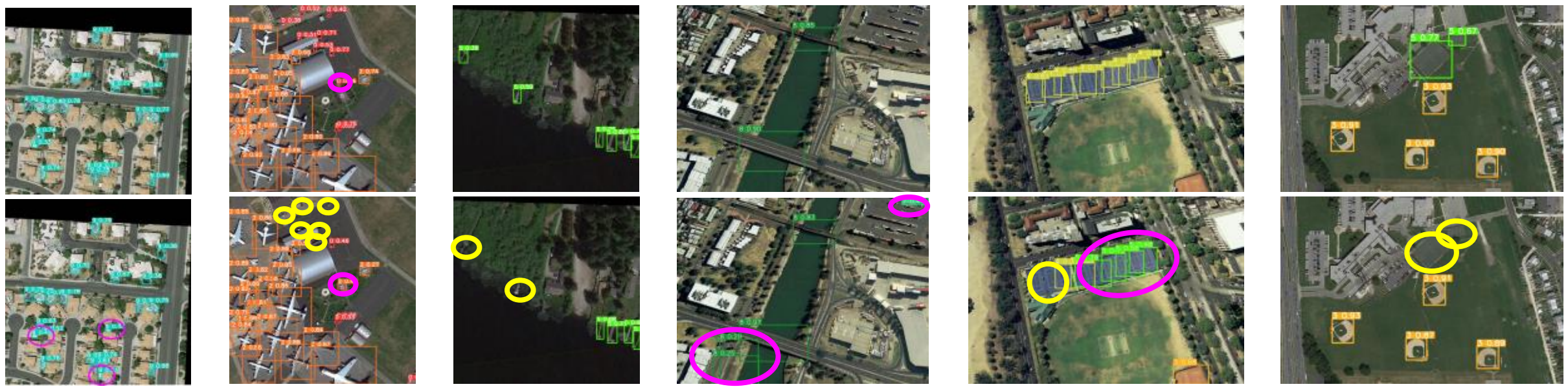

**Fig. 4.** Visualization results. First three columns are DOTA V1.0, last three are NWPU VHR-10. The yellow circle represents missed detection, the purple circle represents wrong detection. The first line is RFWNet, the last line is Baseline

TABLE III
ABLATION EXPERIMENT RESULTS OF HYPERPARAMETERS

| $\gamma$ | 1.0 | 0.9 | 0.8 | 0.7 | 0.6 | 0.5 | 0.0 |
|---|---|---|---|---|---|---|---|
| $\beta$ | 0.0 | 0.1 | 0.2 | 0.3 | 0.4 | 0.5 | 1.0 |
| mAP | 93.7% | **94.6%** | 93.9% | 94.1% | 93.3% | **95.3%** | 94.2% |

TABLE IV
ABLATION EXPERIMENT RESULTS OF IMPROVEMENTS

| Model | mAP/DOTA | mAP/NWPU | Para(M) |
|---|---|---|---|
| Baseline | 69.0% | 88.9% | 7.1 |
| +RFASNet | 72.2% | 92.8% | **5.8** |
| +RFASNet +FBSM | **73.0%** | 93.7% | **6.0** |
| +RFASNet +$L_{WCW}$ | 72.5% | **94.2%** | **5.8** |
| RFWNet | **73.2%** | **95.3%** | **6.0** |

## IV. CONCLUSION

In this study, we propose RFWNet, an efficient and lightweight RSOD algorithm designed to address the challenges of high inter-class similarity, foreground-background imbalance, and small size of objects. RFWNet introduces optimizations in both feature extraction and bounding box regression. Specifically, we developed the RFASNet as the model backbone to capture the context information of remote sensing targets effectively. Secondly, we designed the FBSM to enhance the focus on foreground information while filtering background redundant information. Finally, we designed $L_{WCW}$ as the regression loss function to reduce the model's sensitivity to small target position deviation. The effectiveness of these innovations was validated on two representative datasets. The comprehensive experimental results demonstrate that RFWNet achieves advanced detection results while maintaining an optimal balance between accuracy and computational efficiency.